\documentclass[]{interact}

\usepackage{epstopdf}
\usepackage{subfigure}
\usepackage{color}
\usepackage{ulem}

\usepackage{natbib}
\bibpunct[, ]{(}{)}{,}{a}{}{,}
\renewcommand\bibfont{\fontsize{10}{12}\selectfont}

\theoremstyle{plain}

\theoremstyle{definition}

\theoremstyle{remark}

\begin{document}

\title{Creating Compact Regions of Social Determinants of Health}

\author{
\name{Barrett Lattimer\textsuperscript{a}\thanks{CONTACT Barrett Lattimer. Email: barrettlattimer@sociallydetermined.com} and Alan Lattimer\textsuperscript{a}\thanks{CONTACT Alan Lattimer. Email: alanlattimer@sociallydetermined.com}}
\affil{\textsuperscript{a}Socially Determined, Blacksburg, VA}
}

\maketitle
\begin{abstract}
Regionalization is the act of breaking a dataset into contiguous homogeneous regions that are heterogeneous from each other. Many different algorithms exist for performing regionalization; however, using these algorithms on large real world data sets have only become feasible in terms of compute power in recent years. Very few studies have been done comparing different regionalization methods, and those that do lack analysis in memory, scalability, geographic metrics, and large-scale real-world applications. This study compares state-of-the-art regionalization methods, namely, Agglomerative Clustering, SKATER, REDCAP, AZP, and Max-P-Regions using real world social determinant of health (SDOH) data. The scale of real world SDOH data, up to 1 million data points in this study, not only compares the algorithms over different data sets but provides a stress test for each individual regionalization algorithm, most of which have never been run on such scales previously. We use several new geographic metrics to compare algorithms as well as perform a comparative memory analysis. The prevailing regionalization method is then compared with unconstrained K-Means clustering on their ability to separate real health data in Virginia and Washington DC. 
\end{abstract}

\section{Introduction}
Regionalization is a family of constrained clustering algorithms that define spatially contiguous and homogeneous groups, or regions, in data. Regionalization algorithms have been used in a wide variety of tasks but are especially applicable in grouping geographic communities together. Grouping geographic data is often advantageous due to phenomena such as Tobler's first law of geography, "everything is related to everything else, but near things are more related than distant things" \cite{tobler1970computer}. One such geographically grounded data source is a set of social determinants of health (SDOH). According to Healthy People 2030, SDOH are the conditions in the environments where people are born, live, learn, work, play, worship, and age that affect a wide range of health, functioning, and quality-of-life outcomes and risks. For this paper, we look at community risks in five primary domain areas of SDOH: Economic Climate, Food Landscape, Housing Environment, Health Literacy, and Transportation Network. SDOH have long been considered a good indicator of overall health in communities and in some cases a good predictor of disease or hospital admissions \cite{healthLitHosptial}. This paper aims to quantitatively compare regionalization methods on large scale real world SDOH data and to analyze the advantages and disadvantages of each method.  


SDOH can be defined at a community level and while they are unalterable in the short term, mitigation techniques can be used to improve the SDOH risk of a given community especially in areas of high risk. Healthcare companies and governments alike stand to benefit by identifying more optimized areas of SDOH risk. For example, healthcare companies can use such information to help prevent diabetes by buying food vouchers for areas of high food social risk. Governments could use this information to provide COVID-19 tests to areas with higher financial, housing, or transportation risk. Currently, resources are commonly distributed to communities by zones that were not built to separate communities into common SDOH risk, such as block groups, census tracts, counties, or states. Viewing SDOH through the lens of health agnostic zones can break apart or drown out areas of high SDOH, preventing effective intervention strategies. Ideally a fast and simple clustering algorithm such as K-Means would be used to construct new zones derived directly from SDOH data. However, K-Means is unconstrained and thus is allowed to make geographically muddled zones that would be impossible to take any real world intervention. Providing intervention and relief to high SDOH risk regions is more optimally effective if those regions are compact and geographically contiguous because many mitigation techniques commonly have a geographic radius of impact. Compactness is the measure of density given a region or shape. Contiguity pertains to the connectedness of a region, and in the geographical sense a contiguous region is one in which all elements in a given region must share a border with another element in the same region. In this paper we use regionalization algorithms as a way to automate the creation of geographically contiguous regions of SDOH risk. 

Many regionalization methods have been proposed, the most popular of which are Agglomerative Clustering, REDCAP \cite{REDCAP}, AZP \cite{AZP}, SKATER \cite{SKATER}, max-p-regions \cite{maxp} \cite{maxpHeuristic}, and many others \cite{wan2018dasscan} \cite{geoSOM}. Many of these regionalization methods have been compared to one another \cite{quantCompare} \cite{dao2018detecting} and individually stress tested; however, this has only been done on small scale or simulated data. Additionally, no geographic comparisons between different algorithms generated regions have been performed. Instead, research in evaluating regionalization methods has focused on the effectiveness of clustering on small artificial data sets which results in a lack of testing on large scale real world data. For example, many individual papers evaluating a single regionalization approach will use 1,000 as the maximum number of spatial units. A quantitative comparison was run between some popular regionalization methods in \cite{quantCompare} running tests up to 3,200 spatial units. By far the largest test that has been performed on any regionalization method is REDCAP run on roughly 1 million data points \cite{nlpRegions}. These experiments, however, did not compare regionalization methods to one another nor did it provide any sort of timing or scaling information about REDCAP. The data in this paper compares regionalization methods beginning at 8,000 data points and scales over 10 intervals up to 1 million data points.

We use regionalization to identify high risk regions in communities of real data at scales orders of magnitude higher than has been previously tested. Additionally, our experiments use real data from across the United States that can be further matched with health trends to do real prediction in the future. Our study uses time metrics, which have only been used to compare regionalization methods once at much smaller scales \cite{quantCompare}. Additionally, we include memory comparisons and popular geographic metrics taken from gerrymandering studies, both of which have never been used to compare regionalization methods.


\section{Related Work}

Using regionalization algorithms to create spatially contiguous homogeneous neighborhoods is a long studied problem that has taken the form of many different approaches. There are some comparison studies that exist under the topic of regionalization but only focus on performance within a single algorithm \cite{justREDCAP}. To our knowledge, there exists three previous papers comparing the performance of a large number of regionalization algorithms \cite{quantCompare,dao2018detecting,folchCompare}. All three deal mostly with simulated data sets that focus on quality the of the regionalization algorithms on a relatively small scale. Our paper on the other hand,  stress tests regionalization algorithms on real data in increasing orders of magnitude while monitoring to ensure the quality of results. Our paper is the first of these comparative papers to monitor memory use as well as geographic compactness. The most relevant of the comparative papers to our study is \cite{quantCompare} due to the types of algorithms and metrics used. 

Regionalization methods have previously been tested on case study data \cite{quantCompare, REDCAP, maxpHeuristic, geoSocialData, nlpRegions, guo2018detectingMovement, segUSA, geoSOMClustering}. Most papers, however, only run these case studies on a fixed number of data points \cite{geoSOMClustering, quantCompare}. Furthermore, some papers \cite{segUSA, REDCAP} alternate the number of regions but not the number of data points. Others \cite{nlpRegions} evaluate over both changing numbers of regions and data points but only focus on the performance of a single algorithm. Our paper shows how multiple regionalization algorithms perform against each other when dealing with larger and larger amounts of data which is more relevant to real world applications. Notably, our paper is the only one that compares the performance of a large number of regionalization algorithms over real data.

Regionalization algorithms can be split into two broad camps, spatially implicit and spatially explicit, based off how they treat contiguity within the model \cite{regionDef}. Spatially implicit models take the idea of contiguity within a region more as a strong suggestion than a requirement; a good example of this would be manually weighting a geographic element higher in K-Means to encourage compact clusters. Spatially explicit models, on the other hand, rigorously enforce spatial constraints; a good example of this would be agglomerative clustering using strict restrictions to only merge clusters that are spatially contiguous. Many studies \cite{dao2018detecting, geoSOMClustering, geoSocialData, geoSOM, kolak2020quantification} will use spatially implicit models in their study of regionalization or neighborhood building, however, due to the geographic constraints of SDOH clustering and the aim to create actionable compact zones we take a similar approach as \cite{quantCompare} and restrict our regionalization methods to spatially explicit. Building off of \cite{quantCompare} we only use regionalization algorithms that have been initially shown to be time efficient (REDCAP, SKATER, AZP) as well as adding on a new lightweight baseline (Agglomerative Clustering) and a more mathematically rigorous model (Max-P Heuristic).

There are very few studies on the regionalization of SDOH \cite{kolak2020quantification}. To the best of our knowledge all other studies using regionalization on SDOH have only used spatially implicit regionalization methods without exploring or comparing to any other algorithms. These studies do not acknowledge the downstream benefits of having contiguous and compact regions nor the scalability of the algorithm in use. To the best of our knowledge the performance of spatially explicit regionalization algorithms have never been tested on SDOH data. This paper aims to find the optimal spatially explicit regionalization algorithm to create homogeneous neighborhoods of SDOH across a large variety of scales.

\section{Methods}

\begin{figure}
    \centering
    \includegraphics[scale=.5]{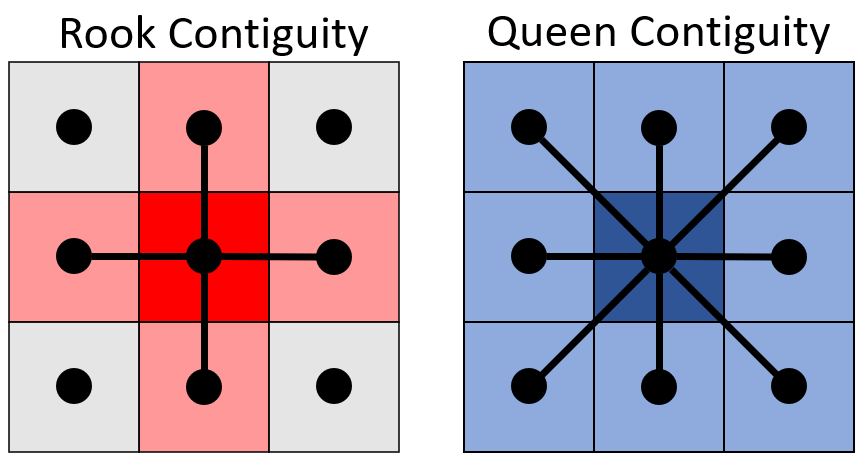}
    \caption{Comparing rook and queen contiguity. Each square is representative of geographic shapes and their centroids represented as black dots. Graphs are generated by treating centroids as vertices and creating edges between vertices based on the existence of borders between the geographic shapes.}
    \label{fig:rookqueen}
\end{figure}

Consider a geographic area tessellated by $n$ hexagons. We represent these hexagons as an undirected graph $G=(V, E)$ where $V=v_1, \ldots, v_n$ is the set of hexagons centroids treated as vertices and $E$ is the set of edges connecting spatially adjacent hexagons. Each vertex $v_i$ has an associated SDOH data vector $u$ with $m$ SDOH data factors of interest, $u_i = d_{1}, \ldots, d_{m}$.  Each edge between two vectors is associated with a weight that measures the distance between the vectors associated data points. In this paper we use Euclidean distance Equation (1) to calculate the distance between two data points. To formalize this notion, let 

\begin{equation}
    d(u, v) = \left(\sum_{k=1}^n (u_k-v_k)^2\right)^{\frac{1}{2}},
    \label{eq:data_distance}
\end{equation}

where $u, v \in D = \langle d_1, \ldots, d_m \rangle \subseteq \mathbb{R}_m$ where $d_1, \ldots, d_m$ are values of SDOH data factors of interest associated with two vertices $v_i, v_j$.

Spatial adjacency can be defined in two main ways, rook and queen contiguity as seen in Figure \ref{fig:rookqueen}. A rook contiguity graph is defined as creating an edge between two data points $v_i, v_j \in G$ if they share a nonzero length geographic border. Queen contiguity is defined as creating an edge between two data points $v_i, v_j \in G$ if they share a nonzero length geographic border or vertex. For this paper we used queen contiguity to define edges in $G$.

In this paper we expand the contiguity of real world SDOH data to span over areas of missing data. SDOH data is initially represented geographically as fixed size tessellated hexagons, however, many gaps exist in non residential areas such as forests or bodies of water. To extend contiguity over these gaps we generate Voronoi diagrams from the geographic centroids of the hexagons. A continuous graph using queen contiguity is then generated over the resulting polygons which we then break into partitions for regionalization. 

A partitioning of $G$ of size $n$ can be defined as $G_1, G_2, \ldots G_n$ such that $G_1 \cup G_2 \cup \ldots \cup G_n = G$ and $G_1 \cap G_2 \cap \ldots \cap G_n = \emptyset$. Let each of the partitions $G_k, k\in (1,n)$ be connected sub-graphs of $G$, so for any pair of nodes $v_i, v_j \in G_k $ there exists a path between them within $G_k$. $G_k$ can also be referred to as a cluster or in our case, a region. This partitioning of $G$ is the formal output of a regionalization algorithm. Regionalization can then be defined as minimizing some objective function over $n$ partitions of $G$. For this paper we minimize the objective function of inter region variability defined as follows in Equation (2). Let $\mu (G_k)$ represent the average SDOH data point value in the partition $G_k$ and $x_i$ a SDOH data point associated with some $v_i \in G_k$. Let 

\begin{equation}
    W(G_1, \ldots, G_n) = \sum_{q=1}^n \sum_{i\in G_{q}} d(x_i, \mu (G_q)).
    \label{eq:within_variability}
\end{equation}

By minimizing the objective function $W(G_1, \ldots, G_n)$ in Equation (2) we find regions that group together similar SDOH data points while still enforcing contiguity. It is also worth noting that between cluster variation $B(G_1, \ldots, G_n)$ defined in Equation (3) shows the distance between $\mu (G_k)$, the average SDOH data point value in the partition $G_k$, and $\mu (G)$ the average SDOH data point value over the entire graph. Let

\begin{equation}
    B(G_1, \ldots, G_n) = \sum_{q=1}^n \sum_{i\in G_{q}} d(\mu (G_q), \mu (G)).
    \label{eq:between_variability}
\end{equation}

A regionalization algorithm then is a way to minimize the objective function such as in Equation (2) while adhering to spatial constraints. The regionalization algorithms used in this study are defined in the following sections. 

\subsection{Agglomerative Clustering}
Hierarchical clustering is a family of clustering algorithms that clusters data by either a top down or bottom up approach. Agglomerative clustering is a specific subset of hierarchical clustering that uses the bottom up approach. Top down approaches begin with the entire data set and recursively splits the data into two further groups according to some optimization function until each data point is in its own cluster. Conversely, bottom up approaches begin with each data point in its own cluster and successively merges data points by some optimization function until there is only one overarching cluster. Commonly data is visualized as a tree called a dendrogram, with the top or root node representing the entire data set in one cluster while the leaves representing each data point as an individual cluster. Travelling from the leaves to the root of the dendrogram, each level represents the merging of two clusters, thus the number of clusters corresponds to the level in the tree. A certain number of clusters can be achieved by cutting the dendrogram at the level with that many clusters. For agglomerative clustering, the algorithm begins at the leaves, with each data point being in their own cluster. Clusters with the "shortest distance apart" are successively merged step by step until the desired number of clusters is reached.

There are four metrics in agglomerative clustering that define the "shortest distance apart" for regions and are used to determine which regions should be merged next. Equations (4-7) respectively define the ward linkage, complete linkage, average linkage, and single linkage distance metrics where $m_i$ represents the centroid of a region $i$ and $n_i$ the number of data points in a given region $i$.  Ward linkage minimizes the sum of squared differences within regions, essentially minimizing intra-cluster variance, Equation (4). Complete linkage minimizes the maximum distance between regions by only observing the two points between regions which are farthest away, Equation (5). Average linkage minimizes the average distance between all possible pairs of data points between two regions, Equation (6). Single linkage, somewhat of an opposite from complete linkage, minimizes the closest distance between regions by only observing the two closest points between regions, Equation (7). This paper uses ward linkage to achieve the goal of forming regions of communities that have similar levels of SDOH. Let

\begin{equation}
    d_{W}(G_i, G_j) = 
    \sum_{q \in G_i \cup G_j} \lvert\lvert x_q - m_{ G_i \cup G_j}\rvert\rvert ^2 - 
    \sum_{q \in G_i} \lvert\lvert x_q - m_{ G_i}\rvert\rvert ^2 - 
    \sum_{q \in G_j} \lvert\lvert x_q - m_{G_j}\rvert\rvert ^2.
\end{equation}
\begin{equation}
    d_{CL}(G_i, G_j) = \max_{u \in G_i, v \in G_j} d(u,v),
\end{equation}
\begin{equation}
    d_{AL}(G_i, G_j) = \frac{1}{n_i}\cdot \frac{1}{n_j} \sum_{k=1}^{n_i}\sum_{l=1}^{n_j} d(u_k, v_l),
    \label{eq:dist_avg_linkage}
\end{equation}
\begin{equation}
    d_{SL}(G_i, G_j) = \min_{u \in G_i, v \in G_j} d(u,v).
\end{equation}

In general performing agglomerative clustering can become very expensive in time and memory because it has to check all possible combinations of regions at every step. When contiguity constraints are enforced however, the possible combinations of regions shrinks dramatically because only connected regions can be merged. This subtle perk of agglomerative clustering has a large impact, because the potential speedup scales with the number of data points used in the clustering.

\subsection{SKATER}
SKATER, which stands for Spatial' K'luster Analysis by Tree Edge Removal, was proposed by \cite{SKATER} and uses the power of minimum spanning trees (MST) to create regions of homogeneous contiguous regions. SKATER fundamentally changes the problem of regionalization into one of optimal graph partitioning. Spanning trees are defined as follows, given our graph $G=(V,E)$, a spanning tree is a subset of the edges $E_0 \subseteq E$ such that all the vertices $v\in V$ are connected together without any cycles. A MST is the spanning tree in which the sum of edge weights is minimum amongst all other possible spanning trees. Creating an MST is proven to be NP-hard so a number of efficient algorithms such as the Prim or the Kruskal algorithm are used in SKATER.  

SKATER begins with a weighted connectivity graph $G=(V,E)$ where $V$ is our set of data points and $E$ is the set of edges. The weight of each edge in $E$ is inversely proportional to the similarity of SDOH data points between the regions it connects. A MST is then formed over $G$ and edges in order of most dissimilar are pruned iteratively until the desired number of regions or some other stopping rule is reached. It is worth noting that removing any edge in an MST defines a further partition of the graph $G$, thus pruning an edge automatically makes a new region.

\subsection{REDCAP}
Improving upon SKATER, REDCAP \cite{REDCAP} defines a family of 6 functions that attempt to improve MST formation by building the MST with agglomerative clustering and then partitioning the MST to obtain regions. Agglomerative clustering is used to iteratively merge regions as discussed in Section 3.1 until only one region exists which by definition is a spanning tree. The resulting spanning tree is then partitioned similar to SKATER in Section 3.2 until the desired number of regions or some other stopping rule is reached. The 6 algorithms that are part of the REDCAP family use either single linkage, complete linkage, or average linkage agglomerative clustering as described in Section 3.1 in combination with either first-order or full-order constraints to calculate the MST. All 6 members of the REDCAP family using the same partitioning function on the MST. Note the algorithms in REDCAP are different from pure agglomerative clustering as there is a further step of creating an MST and pruning based on an objective function such as the one described in Equation (2).

Order constraints are rules that define which edges are used when performing linkage calculations. An edge can be defined as first-order if it connects two spatial regions according to a predefined contiguity matrix. Using a first-order constraint with agglomerative clustering prunes edges as the clustering runs, only keeping edges that directly connect two regions, affecting the inputs to the linkage calculations. Full-order constraint, however, uses all existing edges between two regions when computing linkage. 

For this paper, we used REDCAP with full order complete linkage Equation (5) following previous studies \cite{quantCompare, REDCAP} as well as REDCAP with full order ward linkage Equation (4) to compare against pure agglomerative clustering with ward linkage. 

\subsection{AZP}
AZP (Automatic Zoning Procedure) was originally proposed by \cite{AZP} to supplement the creation of zones using census data via a local optimization approach. AZP begins with $n$ data points and aims to turn them into $m$ regions. The $n$ data points are randomly assigned into $m$ regions while adhering to contiguity constraints. A list of the $m$ regions is formed. A random region $k$ is removed from the list of regions and all data points bordering region $k$ not in region $k$ are put into a set $s$. Data points are then randomly pulled out of set $s$ iteratively and if swapping the given data point to region $k$ gives an improvement in the objective function then the point is assigned to region $k$. This process repeats until the list is exhausted for region $k$, and there are no more possible improvements. Then another region is then randomly selected, and the process repeats until no further improvements can be made. 

There are two popular variations of AZP proposed by \cite{newAZP}, namely AZP-Simulated Annealing and AZP-Tabu. Due to the findings of \cite{quantCompare} revealing the poor timing and performance of these two variations, we used the original version of the AZP algorithm for this paper.

\subsection{Max-P-Regions}
The max-p-regions problem, introduced by \cite{maxp} and further improved by \cite{maxpHeuristic}, is unique amongst the other methods under comparison as it strictly uses a threshold value to determine the number of regions rather than a set number of clusters. Using a set threshold value allows max-p-regions to depend on the data to aggregate into regions rather than an arbitrary number of clusters.  Max-p-regions is broadly composed of three main stages: region growth, enclave assignment, and local search. The first stage, region growth, randomly selects an unassigned seed unit and then adds to its cluster neighboring data points that have not yet been added to a cluster until the minimum threshold is reached or no unassigned neighboring data points can be found. If the region cannot meet the minimum threshold, it is called an enclave and added to the enclave set. Once the data set is broken into regions and enclaves, every enclave is assigned to a neighboring region with the smallest dissimilarity. Lastly, similar to AZP, in local search data points are traded between regions while ensuring the contiguity and threshold constraints are still met until no further improving trades can be made.

Max-p-regions suffers from computational bottlenecks both in memory and run time for large problems due to the algorithm being NP-hard \cite{maxp}. However, due to the popularity and accuracy of this method, we felt it was appropriate to include max-p-regions in this paper for comparison purposes. 

\section{Data and Metrics}

\begin{figure} 
\centering
\resizebox*{.5\textwidth}{!}{\includegraphics{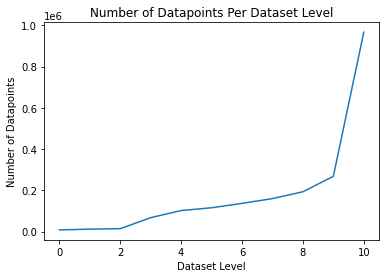}}
\caption{The number of data points in each data set "level". These levels slowly add data points contained in counties expanding out in concentric circles of counties. These levels begin in Washington DC until they eventually consist of all of Virginia and Washington DC} \label{fig:1}
\end{figure}
\subsection{Data}

Geographically, our SDOH data is represented by centroids of  hexagons tessellated across the United States with each hexagon having an associated 5 dimensional SDOH feature vector. The hexagons have a diameter of 200 meters in urban areas and 400 meters in rural areas. Hexagons that are designated as non-residential zones (streets, lakes, etc.) are removed from the data set. The feature vector associated with each hexagon consists of five SDOH metrics: Economic Climate, Food Landscape, Housing Environment, Health Literacy, and Transportation Network. The hexagon-data used in this paper was broken up into 11 geographic "levels" by slowly increasing the number of counties included in Virginia and Washington DC. The geographic levels begin with just Washington DC and expand out in concentric circles of counties into Virginia every level after until the final level, Level 10, which consists of all of Virginia and Washington DC. The rate at which counties are added roughly doubles the amount of hexagons used in the previous level, as shown in Figure \ref{fig:1}.

A number of different metrics are used in the analysis of the six different regionalization methods under comparison. The metrics in use can broadly be split into two categories, unsupervised and geographic. Note because this paper uses real world data and the underlying ground truth regions are not known we are not able to use supervised metrics. 

\subsection{Unsupervised Metrics}
We use unsupervised metrics to measure the two data driven goals of regionalization, homogeneity within regions and heterogeneity between regions. That is, we want each region to represent similar SDOH, but different regions to represent different SDOH. We use multiple metrics to evaluate different aspects of this problem, as shown in the list below.

\begin{itemize}
    \item Calinski-Harabasz index, Equation \eqref{eq:cal_har}: The CHI proposed by \cite{calinski1974dendrite} defines a ratio between within-cluster homogeneity $W(k)$ as defined in Equation \eqref{eq:within_variability} and between-cluster heterogeneity $B(k)$ Equation \eqref{eq:between_variability}. This index can be interpreted as a compactness score of the regionalization clustering when ignoring geography entirely. The CHI will be high for the regionalization we hope to achieve, one that groups highly similar data points in the same region and has large differences in data points between regions. Otherwise, a low CHI is indicative of clusters with dissimilar data points that are not distinct from one another.
    
    \item Silhouette Coefficient, Equation \eqref{eq:sc}: The silhouette coefficient (SC) is defined as a ratio between the average intra-cluster distance $a$ and the average inter-cluster distance $b$. Put another way, $a$ is the average distance between points within a cluster and $b$ is the average distance between clusters. Since we measure the distance between data vectors as in \eqref{eq:data_distance}, SC helps us interpret how well our regions classify the data. The SC ranges from -1 to 1, where an SC near 1 means data points are well matched to their cluster and poorly matched to other clusters, while an SC near -1 means the clusters do not match up with its data points well.
    
    \item Average number of high risk domains (AHR): We define the number of high risk domains in a single SDOH data point as the number of SDOH data factors of interest that are in the top 30\% nationally amongst the other SDOH data factors of interest of its kind. The top 30\% can be determined by binning each SDOH data factor of interest into bins 1 through 5 nationally and treating the 4 and 5 bins as high risk. AHR simply averages the number of high risk domains per data point in the region. A high AHR represents a higher risk region while a low AHR represents a low risk region, providing insight if clusters are separating high and low risk areas well. 
    
    \item Sum-Squared Errors, Equation \eqref{eq:sse}: Measures inter region variability by taking the sum of the squared difference between all points in a region. The higher the SSE is, the more variability (more error) there is between SDOH data points in a region. We normalize SSE in this paper by dividing by the number of points per region to compare regions of different sizes and across different data levels.  
\end{itemize}

\begin{equation}
    CHI(k)=\frac{k-n}{n-1}\cdot \frac{B(k)}{W(k)}
    \label{eq:cal_har}
\end{equation}
\begin{equation}
    SC = \frac{b-a}{\max(a, b)}
    \label{eq:sc}
\end{equation}
\begin{equation}
    SSE = \sum_{i=1}^n \sum_{j=1}^n (x_i-x_j)^2
    \label{eq:sse}
\end{equation}

\subsection{Geographic Metrics}
Geographic metrics are applied to evaluate how geographically compact and actionable the generated regions are. As discussed previously, geographically compact and contiguous regions are important to regionalizing SDOH, however, this does not guarantee our regions will be compact and uniform. Taking inspiration from gerrymandering studies \cite{sun2021developing} multiple different types of metrics were used to measure the quality of each regionalization methods geographic compactness. The metrics used are listed below. 

\begin{itemize}
    \item Percent Overlap, Equation \eqref{eq:percent_overlap}: In this metric we fit a concave hull $\alpha$, or alpha shape, to each region $r$ to get a smoothed outline of each generated region. We then calculate the pairwise percentage overlap between all regions and average these percentages. This metric is used as a proxy to estimate how well a regionalization method partitions the given data in a non overlapping manner with well defined borders.
    
    \item Convex Hull, Equation \eqref{eq:convex_hull}: The convex hull (CH) metric is the ratio between the area of a region $A_R$ and the area of the minimum convex polygon that can enclose the region $A_P$, also known as its convex hull. This ratio is between 0 and 1 where 1 signifies a more compact region. 
    
    \item Polsby-Popper, Equation \eqref{eq:pp}: The Polsby-Popper (PP) metric proposed by  \cite{polsby1991third} is calculated by taking the ratio between the area of a region $A_R$ and the area of a circle who's circumference is equal to the perimeter of the district $A_C$. The PP score is between 0 and 1 where a score of 1 represents a more compact district. 

\end{itemize}

\begin{equation}
    PO = \frac{1}{n^2-n}\cdot \left(\sum_{i=1}^n \sum_{j=1}^n \frac{area(\alpha (r_i) \cap \alpha (r_j))}{area(\alpha (r_i))} - \sum_{i=1}^n \frac{area(\alpha (r_i) \cap \alpha (r_i))}{area(\alpha (r_i))}\right)
    \label{eq:percent_overlap}
\end{equation}
\begin{equation}
    CH = \frac{A_R}{A_P}
    \label{eq:convex_hull}
\end{equation}
\begin{equation}
    PP = 4\pi \cdot \frac{A_R}{A_C^2}
    \label{eq:pp}
\end{equation}

During all our experiments we set the number of clusters to 5, as previous exploratory data analysis showed this was a good choice for our data at multiple geographic levels. The SDOH feature vectors are normalized via min-max scaling for each level before regionalization is performed. Queen contiguity is used throughout all experiments to generate connections between data points. Generated regions are labeled 1 through 5 based on their AHR metric with 1 being the lowest AHR (risk) region and 5 being the highest AHR (risk) region. A time limit of 1 hour was put on regionalization methods when running.

\section{Results}
We begin by comparing the regionalization algorithms under Unsupervised metrics and then Geographic metrics in Section 5.1 and 5.2 respectively. Section 5.3 explores Washington DC as a case study on the county level while Section 5.4 runs agglomerative clustering on the largest level in our dataset, Level 10 consisting of Virginia and Washington DC. Lastly, Section 5.5 compares unconstrained clustering and spatially explicit regionalization methods through the lens of real health metrics.

All methods under comparison except agglomerative clustering reached a termination point prior to the final Level 10. REDCAP with complete linkage, REDCAP with ward linkage, SKATER, and AZP all exceeded the working memory on the running machine, 32 GB. Max-P-Regions was only able to complete Level 0 (Washington DC) regionalization within the hour time frame so it is not included in any of the unsupervised or geographic metric analysis. 

\subsection{Unsupervised Results}
\begin{figure}[htb]
\centering
\resizebox*{\textwidth}{!}{\includegraphics{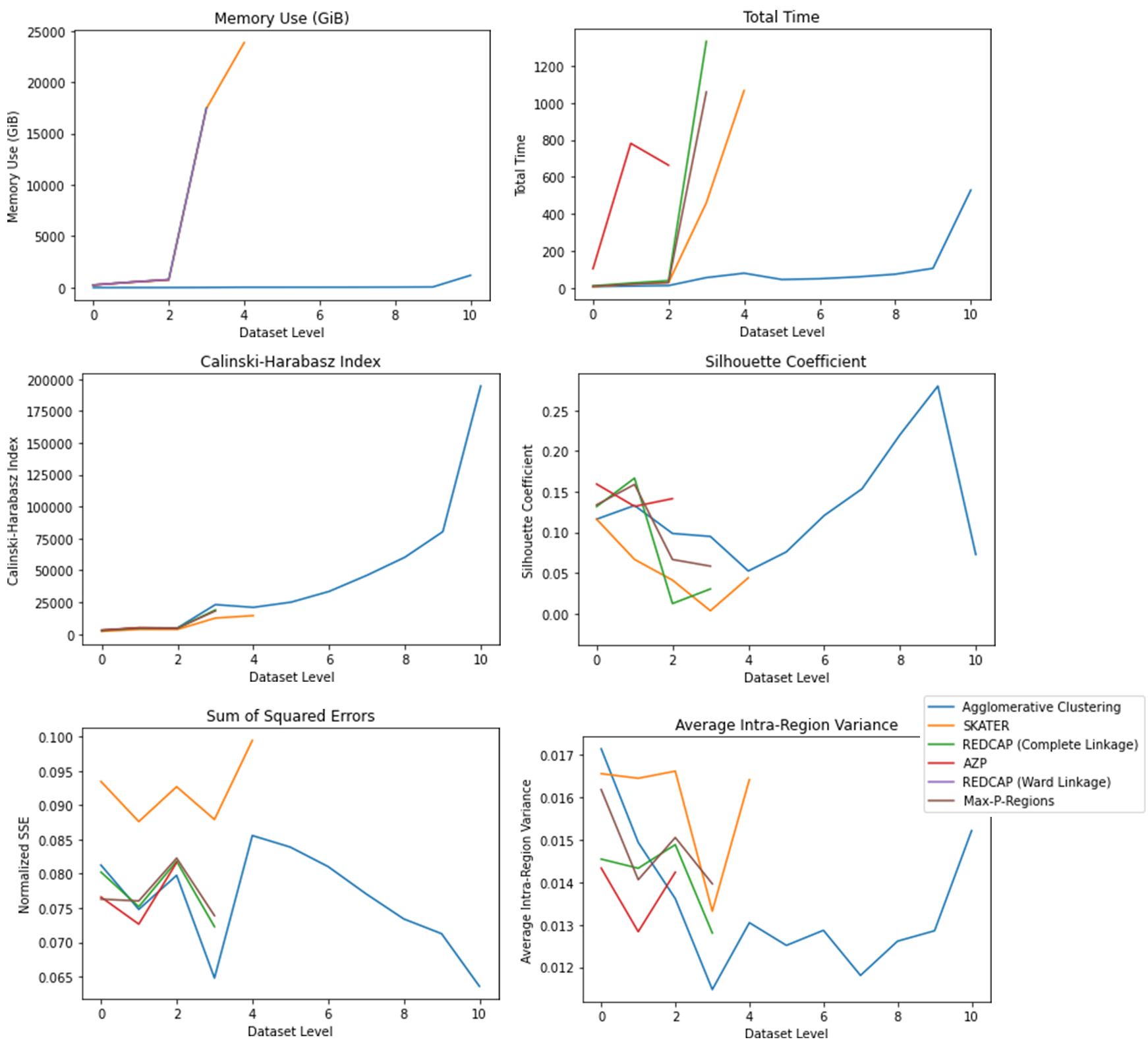}}
\caption{Six unsupervised metrics used to compare the ability of each regionalization algorithm to separate data well with efficiency in time and memory.} \label{fig2}
\end{figure}

To compare each regionalization model's ability to create regions that are homogeneous yet heterogeneous from one another, a number of unsupervised metrics were recorded as shown in Figure \ref{fig2}. AZP exceeded the memory limit at Level 3, the REDCAP model family at Level 4, and SKATER at Level 5. The memory peak for each regionalization method also coincides with a large spike in time for clustering as shown in Figure \ref{fig2}. Differing from other algorithms, AZP begins with a large amount of time to cluster and stays that way until it runs out of memory. When the number of data points increases from 14,000 in Level 2 to 67,000 at Level 3, the REDCAP family and SKATER both hit a critical point of increase in time to cluster and memory. After Level 2 the time to cluster for the REDCAP family and SKATER becomes highly nonlinear and quickly exceeds our computational limitations. Agglomerative clustering maintains an almost constant time to cluster over all levels, making it the superior method in terms of time and memory. 

The silhouette score in Figure \ref{fig2} shows agglomerative clustering starting off lower than the REDCAP family and AZP. AZP remains superior in this metric until it runs out of memory. The REDCAP family drops off with SKATER by Level 2 well below agglomerative clustering. Agglomerative clustering notably has an inflection point at Level 4, constantly increasing afterwards and eventually achieving a higher silhouette score than all other models. Interestingly a similar trend is seen in the CHI in Figure \ref{fig2}. SKATER again performs the worst while agglomerative clustering sees a large increase after an inflection point at Level 4. These two metrics most notably point out flaws in SKATER's clustering ability and that the quality of agglomerative clustering increases with the size of data after an inflection point. 

The SSE metric in Figure \ref{fig2} shows a similar trend in error between all methods except SKATER which had significantly more error. Level 4 is again problematic for agglomerative clustering but acts as an inflection point that leads to much lower levels of error as the levels increase in size. 

Intra-cluster variance which each clustering method aims to minimize in Figure \ref{fig2} shows increasingly better clustering as the size of the data increases as well as some similar trends as all the other unsupervised metrics. SKATER again performs the worst out of all the metrics, but interestingly still outperforms agglomerative clustering in the smallest data set level. AZP again outperforms all the other models before it runs out of memory. Agglomerative clustering levels out as the data set sizes increase, again showing that as the data size increases its performance increases. 

The unsupervised metrics tell an interesting story of how the quality of each regionalization method scale over larger data sets. AZP generally generates the best results for small data sets but very quickly runs out of memory once the level sizes get larger. SKATER has the worst performance across all the unsupervised metrics in general. Agglomerative clustering outlasts all other regionalization models, requiring essentially constant memory, almost constant time, and increases the quality of its clustering as the data size increases. 

\subsection{Geographic Results}

\begin{figure}[htb]
\centering
\resizebox*{\textwidth}{!}{\includegraphics{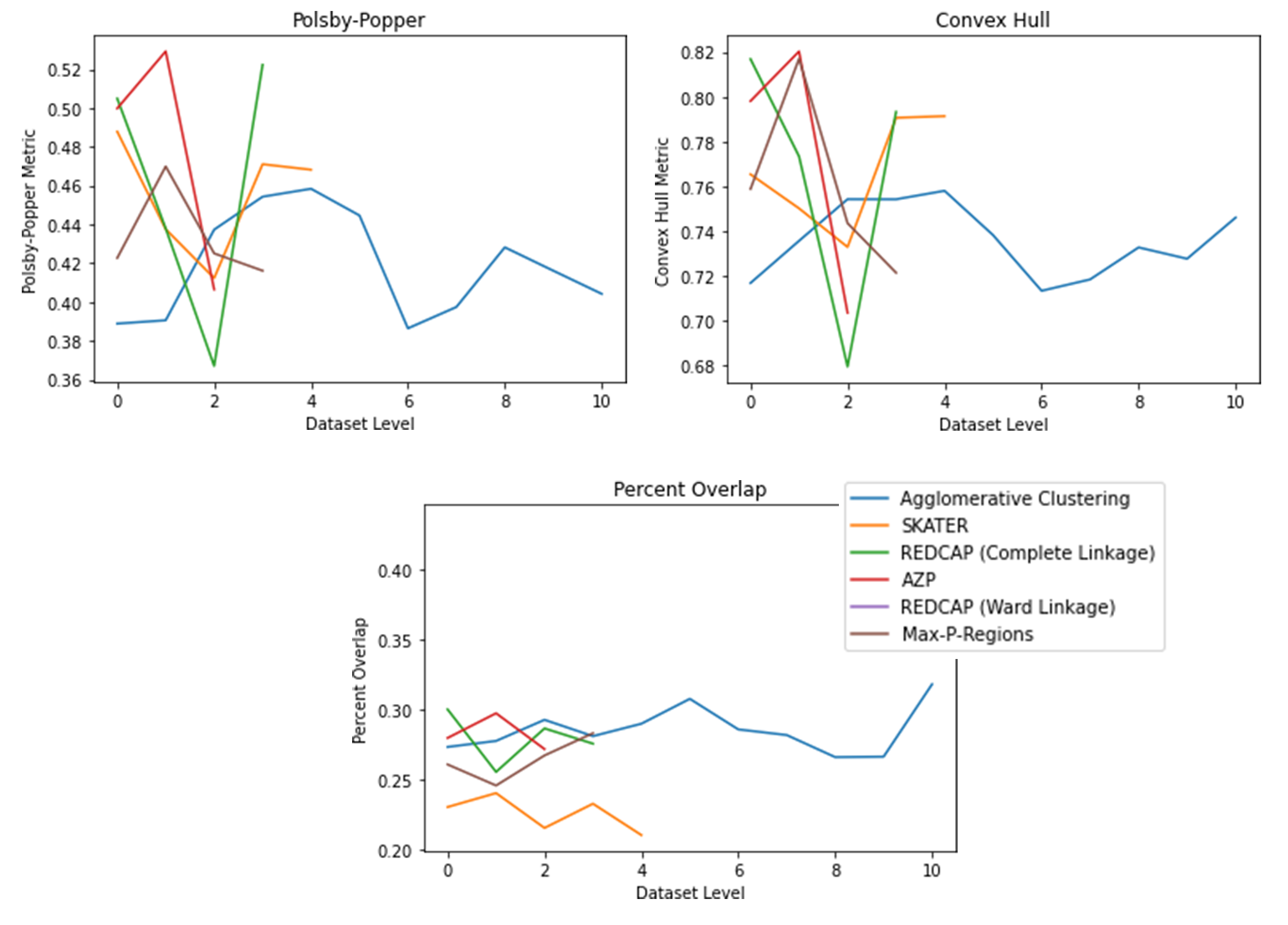}}
\caption{Three metrics relating to the geography of generated regions of all algorithms under comparison. The Polsby-Popper and Convex-Hull metrics measure the geometric compactness of generated regions. Percent overlap shows the ability of regionalization algorithms to partition the data set into non-overlapping regions with smooth definite borders.} \label{fig3}
\end{figure}

Taking inspiration from gerrymandering studies \cite{sun2021developing}, we also use multiple geographic metrics to evaluate the compactness and partitioning ability of the regions that are generated by each regionalization method in Figure \ref{fig3}. 

The Polsby-Popper (PP) metric in Figure \ref{fig3} and Convex Hull (CH) metrics in Figure \ref{fig3} tell consistent stories of compactness. Agglomerative clustering, while it was superior in the unsupervised metrics, shows generally the lowest compactness out of all the regionalization methods in all the lower levels other than Level 2. Similar to the unsupervised metrics, an inflection point for agglomerative clustering is seen at Level 4 in both CH and PP metrics. Dissimilar to the unsupervised metrics, however, after agglomerative clustering's Level 4 inflection point the PP and CH metrics show increasingly worse compactness. This could imply that as the amount of data increases, agglomerative clustering will trade geographic compactness for better clustering. AZP is the best geographic compactness option for smaller data sets until it runs out of memory, just like in the unsupervised metrics. SKATER and complete linkage REDCAP provide the most compact regions for more moderate sized data before they run out of memory as well. Notably, REDCAP with complete linkage and AZP score higher relative to other models in PP rather than CH, implying their clusters are more circular in nature than polygons. 

The percent overlap between concave hulls of regions helps determine the partitioning ability of each regionalization method as well as its ability to make well defined borders. We would like our regions to divide a given area into clean partitions such that when the borders are fit to concave hulls, essentially outlining the region, the regions have little overlap. Clean partitions prevent long snaking regions that make intervention more difficult. Following trends set in the PP and CH metrics, percent overlap in Figure \ref{fig3} shows SKATER performing the best while agglomerative clustering is generally performing the worst. Contrary to the PP and CH metrics, AZP and REDCAP complete linkage have poor percent overlap. AZP and REDCAP complete linkage's poor percent overlap performance could be linked to the same reason they both performed worse in the CH metric when compared to the PP metric, because they are not modeled well by polygons but rather by circles. REDCAP ward linkage performs fairly well under this metric initially, showing it may perform well in smaller data set sizes. Agglomerative clustering again experiences an inflection point around Level 4 and 5, however this one is good and leads to a decrease in percent overlap. Agglomerative clustering's inflection point may be due to larger data sets having larger borders and thus a fitted concave hull is far less likely to have sharp overlapping edges. 

The geographic metrics show that agglomerative clustering generally has lower compactness among the models. We also see that for smaller data, AZP is the superior method for generating compact regions. SKATER performs comparatively better across all metrics, especially as the data sets get larger. The REDCAP ward linkage performs well across all the metrics but reduces as the data set size gets larger, while REDCAP with complete linkage experiences dramatic variability as data sizes increase. 

\subsection{Washington DC - Level 0}

\begin{figure}[htb] 

\includegraphics[width=\linewidth]{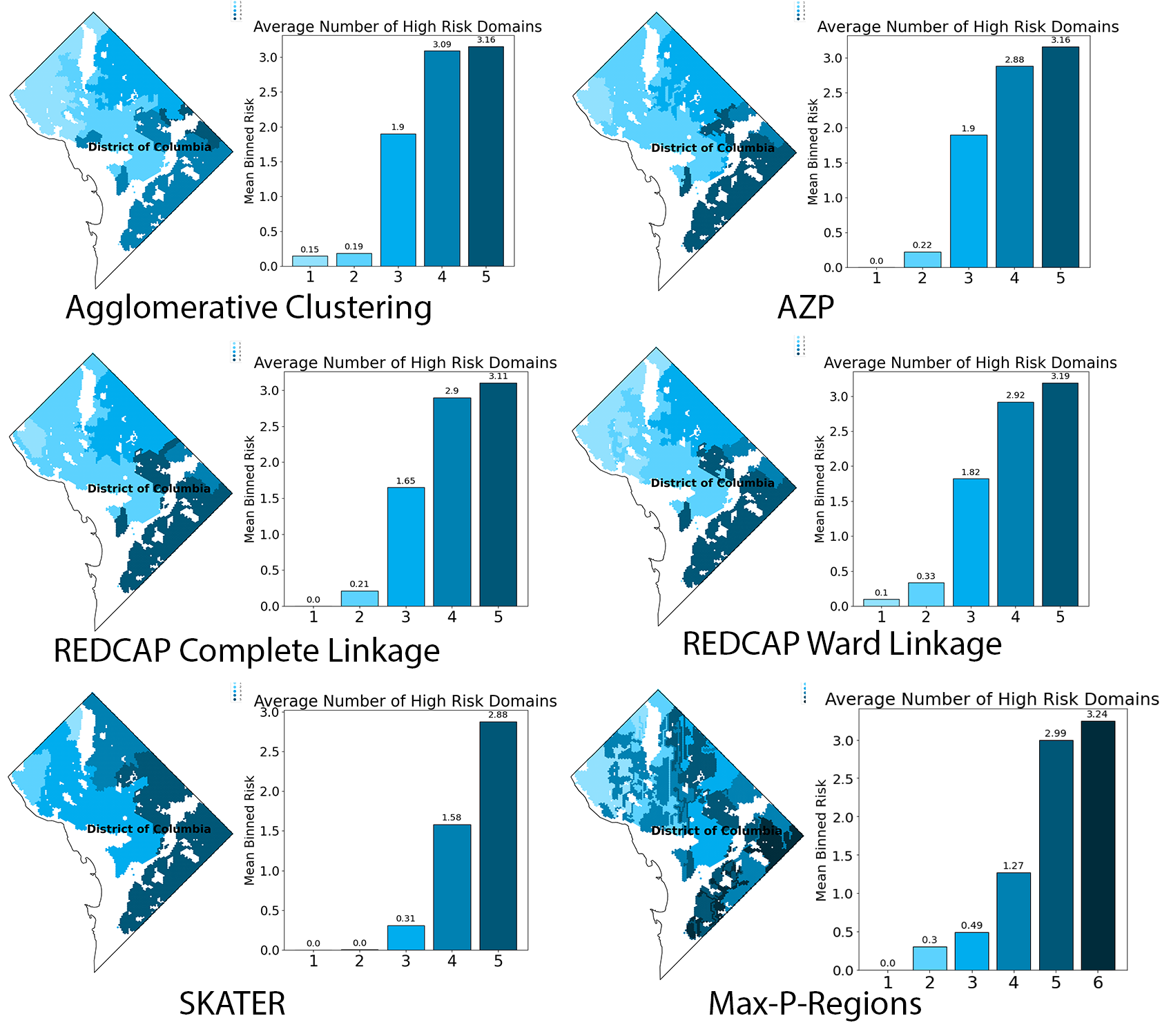}

\caption{Regionalization of Washington DC (data set Level 0) by all regionalization algorithms under comparison. Each geography graph is paired with the average number of high risk domains per data point in each region.} \label{fig:4}
\end{figure}

Level 0, or just Washington DC, presents a good way to visualize some of the differences we outlined in the previous unsupervised and geographic compactness sections, as well as see some real world results using each of the algorithms under comparison. Washington DC consists of 8,137 data points, with white spaces representing non residential areas. Linkage is not restricted from stretching over non residential areas up to a given distance. All methods are set to generate 5 regions and max-p had a minimum threshold of 10 percent of points per region, resulting in 6 regions. 

The smoothness of SKATER's partitioning of Washington DC in Figure \ref{fig:4}, with 1 as the lowest risk region and 5 at the highest risk region, makes it easy to see how SKATER got such superior scores in the geographic metrics, specifically the percent overlap metric. By using pure MST cuts, the very nature of the SKATER algorithm forces smooth borders with low overlap between regions rather than other algorithms such as AZP or agglomerative clustering that have less compact borders. On the other hand, the SKATER AHR distribution, or "risk" distribution, between clusters shows very low variance between the bottom three clusters, explaining the poor performance of SKATER in unsupervised metrics when compared to other methods.

Figure \ref{fig:4} lets us finally see why we wouldn't want to use max-p-regions even if it was feasible in time and space. The regions that created by max-p-regions are spindly, overlapping, and not compact whatsoever. The max-p-regions algorithm stretches the definition of contiguous to its very limits. Max-P-Regions sacrifices any sort of geographic compactness for finding a more optimal solution. 

While each algorithm produces distinct results, there are some overarching consistencies in the regions generated by the algorithms under comparison. For example, each method identifies the south and south-east part of Washington DC as higher risk while picking out the north-west part as low risk. Commonalities between results show that even though different regions are generated, there is a general consensus among methods of roughly where low vs high risk neighborhoods are located geographically and validates the ability to geographically cluster SDOH data.

\subsection{Virginia}
\begin{figure*} 
\includegraphics[width=\linewidth]{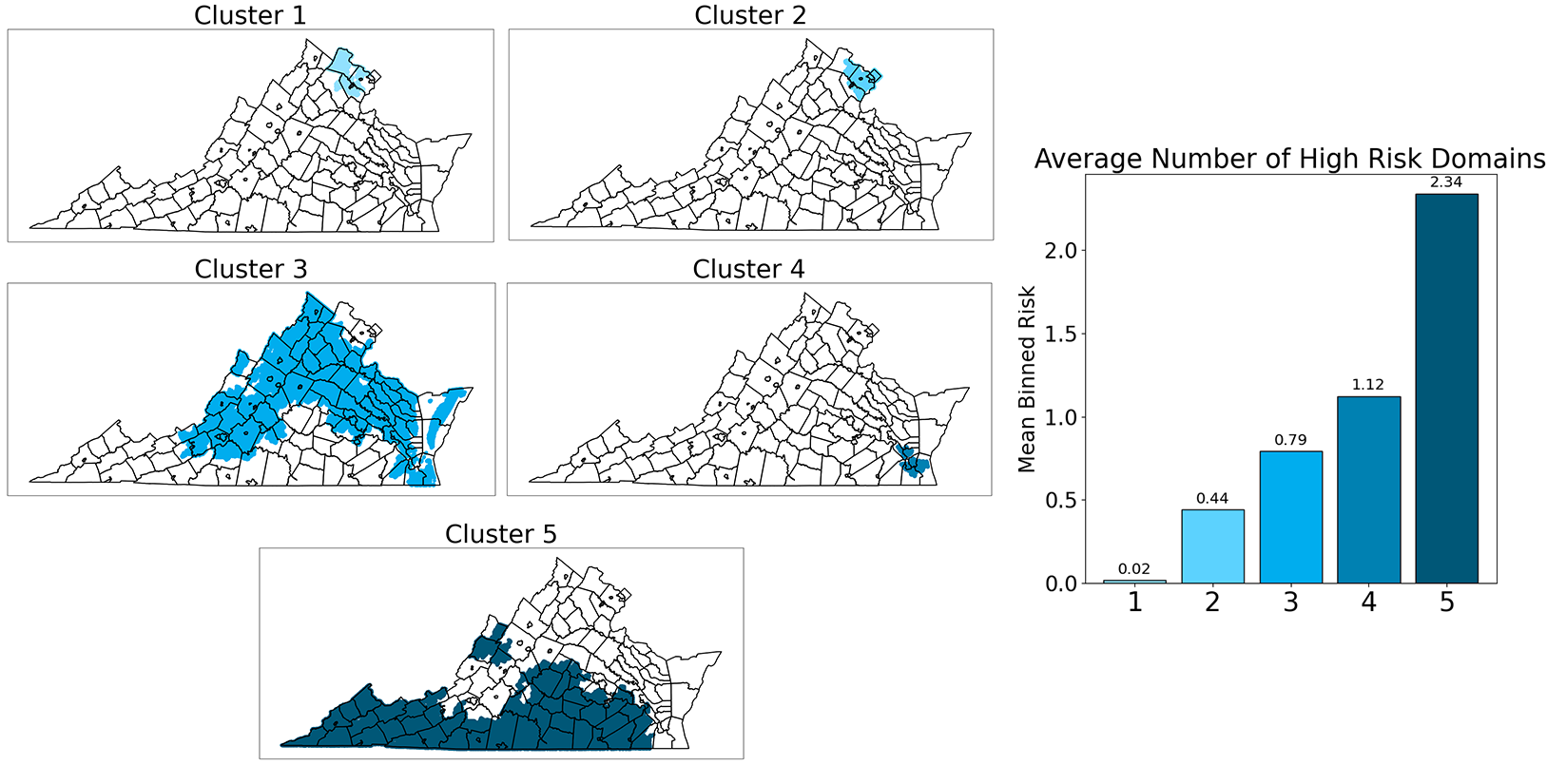}
\caption{Agglomerative clustering run on SDOH in the state of Virginia and Washington DC (data set Level 10). Each region is shown individually mapped on Virginia and Washington DC. Regions are ordered in increasing amounts of high risk domains per data point on average as shown on the right.} \label{fig:5}
\end{figure*}

The largest run, Level 10, which consists of Virginia and Washington DC was only able to be run by agglomerative clustering and the results are shown in Figure \ref{fig:5}. 

Notice that the neighborhoods of risk pay little to no attention to actual county lines. You can see this especially along the border of the high risk region, Region 5, where many counties are split in half. Additionally, notice that these groups of risk aren't influenced by the physical geography of Virginia, in fact both Region 3 and 5 span across all geographic regions of Virginia while showing large differences in risk.

Agglomerative clustering shows the ability in Figure \ref{fig:5} to pick out unique hot spots of risk even when the region size is relatively small. Washington DC and Northern Virginia, certainly the most wealthy part of Virginia and Washington DC, is singled out in the two lowest risk regions. On the opposite end of the spectrum a condensed part of Virginia Beach is identified as having uniquely high risk for a small area in region 4. 

\subsection{Unconstrained Clustering Comparison}
\begin{figure*} 
\includegraphics[width=\linewidth]{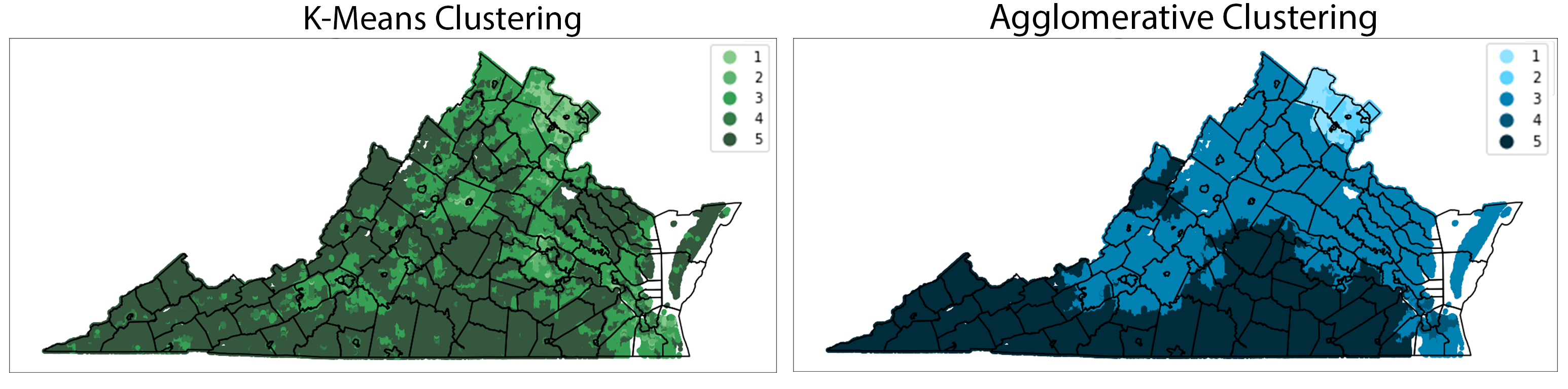}
\caption{Side-by-side comparison between K-Means and agglomerative clustering run on SDOH in the state of Virginia and Washington DC (data set Level 10).} \label{fig:6}
\end{figure*}

This study focuses on how spatially explicit regionalization algorithms compare to one another, but how does the spatial constraints of regionalization inhibit its original goal, forming homogeneous clusters that are heterogeneous from one another. To gauge the change in cluster quality we compare agglomerative clustering to a completely unconstrained and unaltered K-Means algorithm both run on SDOH in the state of Virginia and Washington DC, also known as data set Level 10. Agglomerative clustering is used here as a representative algorithm of the regionalization algorithms under comparison so the analysis could be run on large scale data. Additionally, to compare the quality of regions or clusters formed, a number of health metrics are analyzed for both K-Means and agglomerative clustering. Namely, these metrics are the 2021 Age-Adjusted Suicide Rate, Life Expectancy, and Percentage of Adults with Diabetes in Virginia counties and Washington DC taken from https://www.countyhealthrankings.org/. 

The side by side geographical comparison between agglomerative clustering and K-Means can be seen in Figure \ref{fig:6}. Note that there is very little in common in terms of how K-Means and agglomerative clustering cluster the data other than the DC area being identified as the lowest risk area in the entire geography. K-Means clusters as expected pay little attention to geography whatsoever with clusters fairly evenly distributed throughout Virginia. The lack of compactness in the K-Means clusters is a severe disadvantage in the context of health risk because you no longer have actionable intervention zones in the broader context of the whole state. Both K-Means and agglomerative clustering pay little to no attention to county lines in their cluster assignments.

\begin{figure}[htb]
\includegraphics[width=\linewidth]{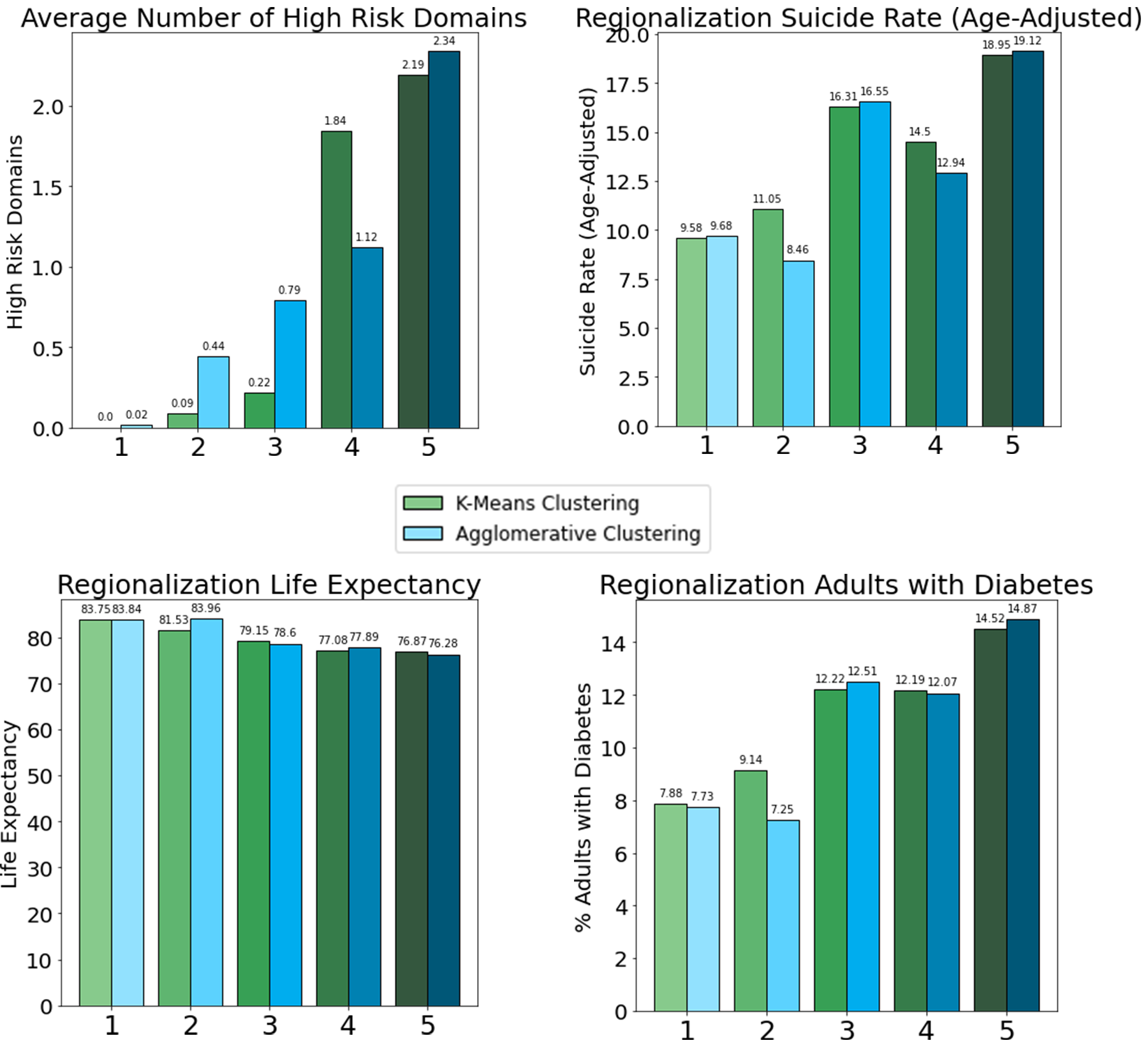}
\caption{Comparing different health metrics between unconstrained K-Means clustering and spatially constrained agglomerative clustering regionalizaiton. Additionally, the number of average number of high risk domains per cluster is shown.} \label{fig:7}
\end{figure}

To further compare unconstrained clustering methods such as K-Means and spatially explicit regionalization methods such as agglomerative clustering we explore how well each clustering correlates to health data. We obtained county level health data for Virginia and Washington DC from countyhealthrankings.org. Each data point was assigned the health value corresponding to their county, and the data shown in Figure \ref{fig:7} is the result of averaging each one of these health metrics per region for each algorithm. For all three health metrics the high risk cluster identified by agglomerative clustering, Cluster 5, has slightly worse health than those identified by K-Means. Both algorithms provide significant levels of separation between different health metrics with the low risk clusters settling on almost equivalent values. The average number of high risk domains in Figure \ref{fig:7} show a more gradual increase for agglomerative clustering and eventually arrive at a higher risk factor than K-Means which shows a large jump in risk between Clusters 3 and 4. This shows a slight advantage in correlation to health metrics for agglomerative clustering while also having the advantage of creating actionable intervention regions. 

The ability of SDOH to separate communities into high and low risk groups is  reflected in the results shown in Figure \ref{fig:7}. Between low and high clusters, life expectancy decreases by roughly 9 percent, the percentage of adults with diabetes increases 92 percent, and the age-adjusted suicide rate increases by 97 percent.

\section{Discussion}
Our results show that agglomerative clustering, AZP, and SKATER perform the best out of all the regionalization methods under comparison. Agglomerative clustering proves to be superior in time, memory, and unsupervised clustering metrics in general when your aim is to process large amounts of data quickly. AZP shows that if you have smaller (around 10,000 points) data sets and a little extra time it's capable of performing the best overall clustering when factoring in unsupervised and geographic compactness metrics. Finally, SKATER is the best method if you wish to find the most geographically compact and smooth regions faster and for larger data sets than AZP can handle, if you're willing to trade for worse performance on unsupervised metrics. All around, for the purposes of regionalization of SDOH, agglomerative clustering is the best method to use due to time and memory scaling superiority, superior unsupervised metric performance, and comparable geometric compactness. 

The comparative nature of this study allows us to ignore the option of simply scaling up our compute power until each method is able to run all the way to Level 10. Without finding extraordinary computing power we are able to log the inflection points of each algorithm and when each method starts to use memory and time at a rate that is no longer realistic or worth the extra cost of computation. We discovered breaking points in algorithms without having to use an unreasonable amount of computing power.

This paper uses the python library pygeoda for all algorithms except for agglomerative clustering which is from sklearn also in python. Pygeoda is ported over from the popular GeoDa software package written in C++. Sklearn and pygeoda are the best libraries available for each respective method available in python, however it seems pygeoda may be less tailored to run in python as it is ported over from C++. Because of this, the weaknesses of some methods may be accentuated when run on large sets of data. We do not attribute the main differences in cluster time or memory to the difference in python libraries because of the large amounts of expensive processing (MST cutting or region neighbor trading) necessitated by all algorithms other than agglomerative clustering.

The assumption that each regionalization method should generate 5 regions at every granularity may have impacted both geographic and unsupervised metrics. For some levels, 5 regions may have actually been ideal, providing quality separation between data points and achieving good clustering while also being able to pick out the correct number of unique contiguous neighborhoods of risk. However, issues of both geographic and clustering significance may have arisen from assuming 5 regions throughout the entire experiment. Given more regions, identified regions could have been subdivided further into a high and low risk neighborhood, increasing the clustering quality and further identifying areas of higher risk. Increasing the number of regions could also start to break off regions that are far too small and statistically insignificant. Given less regions, smaller unique areas of higher risk could be lost in a larger region.

One exciting repercussion of the slow growth rate of agglomerative clustering's time to cluster is the ability to dynamically do regionalization of any given subset of data points in the entire country. Given a precomputed contiguity matrix, which is feasible to store on the back end of a server, regionalization can be done on roughly 250,000 data points in about a minute an a half, and roughly 1 million data points in about 9 minutes on a standard 8 vCPU with 32 GB of memory. Because each data point is a hexagon, we assume the number of neighbors across the United States per hexagon is fairly similar, meaning we can treat these timing measurements as a good estimation of what to expect from regionalization in other parts of the country at a similar scale. The speed of agglomerative clustering opens the door for a completely dynamic regionalization experience with regionalization in under 2 minutes for data set selections under 300,000 data points running on a standard computer.

Especially shown by agglomerative clustering, regionalization even on the multivariate scale is able to separate SDOH effectively. The geographic constraints do not prevent significantly different clusters from forming in vastly different amounts of data. This is important not only for finding communities with high SDOH risk but also any other potentially geographically influenced data field. The large significance of down the road health repercussions such as life expectancy, diabetes, and suicide rates all show significant separation under SDOH clusters as well. The effectiveness of separation opens doors for regionalization to be used to identify high risk areas for real world applications such as intervention zones.

\section{Future Works}
Multiple continuations could be stemmed from this study to further explore how regionalization methods perform on SDOH. Different numbers of clusters could be experimented with to identify how the number of clusters affects regionalization, how the optimal number of clusters scales with data size, and how to find what the optimal number of clusters for any given regionalization method/data size is. Another direction could be running the given regionalization methods with large amounts of compute to see how the methods compare in unsupervised and geographic metrics with higher data levels. Additionally to assist in this effort, new implementations could be written for all of the given methods in python to allow them full utilization of the language and potentially providing a time and memory reduction. Finally spatially implicit regionalization methods or just normal clustering methods could also be included in the comparison to show how these clustering algorithms compare to the spatially explicit methods under comparison in this paper. 

Machine learning methods could be applied to the generated regions using them as abstract zones. Using regions as opposed to low level hexagons could reduce noise and boost the results found by machine learning methods. Results from machine learning models would then have actionable zones that can benefit from real world intervention rather than the much smaller hexagons. Intervention studies using the generated regions could be performed as well to show if intervention in a high risk region benefits the overall risk of the regionalization cluster more than anywhere else. More exploratory data analysis could also be performed on the identified regions to study the distributions of specific SDOH (Food Landscape, Economic Climate, ...) within each region and show how they differ across algorithms, number of clusters, and data set sizes. 

\section{Conclusion}
This paper compared the quality of 6 different regionalization algorithms using multiple new metrics over real world SDOH data at scales orders of magnitude larger than ever tested previously. Unsupervised metrics, geographic compactness metrics, time, and memory were used to evaluate and compare the quality of regionalization between algorithms. We noted the subtle similarities and differences between regions generated by the different regionalization algorithms when applied to our smallest data set size, Washington DC. Notably, agglomerative clustering was the only regionalization algorithm capable of running all 11 data set sizes and was superior in long term unsupervised metrics performance. AZP and SKATER both proved to be good alternative options for small and medium sized data sets respectively. Ultimately for large scale regionalization tasks, agglomerative clustering works the best for multivariate SDOH at scale. We show that SDOH can be clustered well with geographic constraints opening the door for future machine learning work on dynamic SDOH regions.

\bibliographystyle{tfv}
\bibliography{custom}
\end{document}